\documentclass[letterpaper, 10 pt, conference]{ieeeconf}  %

\IEEEoverridecommandlockouts                              %

\usepackage{graphics} %
\usepackage{epsfig} %
\usepackage{mathptmx} %
\usepackage{times} %
\usepackage{amsmath} %
\usepackage{amssymb}  %
\usepackage{algorithm}
\usepackage{algpseudocode}
\usepackage[table]{xcolor} %
\usepackage{float} %
\usepackage{graphicx} %
\usepackage{subcaption} %
\usepackage{wrapfig}
\usepackage{cite} %
\usepackage{lipsum}  
\usepackage{xspace}
\usepackage[colorlinks=true, urlcolor=blue, linkcolor=red]{hyperref}
\usepackage{url}
\usepackage{listings}
\usepackage{xcolor}
\usepackage[final]{changes}

\definecolor{codegreen}{rgb}{0,0.6,0}
\definecolor{codegray}{rgb}{0.5,0.5,0.5}
\definecolor{codepurple}{rgb}{0.58,0,0.82}
\definecolor{backcolour}{rgb}{0.95,0.95,0.92}

\lstdefinestyle{mystyle}{
    backgroundcolor=\color{backcolour},   
    commentstyle=\color{codegreen},
    keywordstyle=\color{magenta},
    numberstyle=\tiny\color{codegray},
    stringstyle=\color{codepurple},
    basicstyle=\ttfamily\scriptsize, %
    breakatwhitespace=false,         
    breaklines=true,                 
    captionpos=b,                    
    keepspaces=true,                 
    numbers=none,                
    numbersep=5pt,                  
    showspaces=false,                
    showstringspaces=false,
    showtabs=false,                  
    tabsize=1,
}

\newcommand{\cmt}[1]{}

\long\def\ignorethis#1{}

\newcommand{\vc}[1]{\ensuremath{\mathbf{#1}}}

\newcommand{\st}{\ensuremath{\vc{s}_{t}}}

\newcommand{\stp}{\ensuremath{\vc{s}_{t+1}}}
\newcommand{\at}{\ensuremath{\vc{a}_{t}}}

\newcommand{\pctab}{\hspace{0.2in}}

\title{\LARGE \bf
Learning a High-quality Robotic Wiping Policy Using Systematic Reward Analysis and Visual-Language Model Based Curriculum
}

\author{Yihong Liu$^{1}$ Dongyeop Kang$^{2}$ Sehoon Ha$^{1}$%
\thanks{$^{1}$YL and SH are with Georgia Institute of Technology, Atlanta, GA, USA
        {\tt\small \{yliu3518,sehoonha\}@gatech.edu}}%
\thanks{$^{2}$DK is with Electronics and Telecommunications Research Institute, Daegu, Korea
        {\tt\small kang@etri.re.kr}}%
}

\begin{document}

\maketitle
\thispagestyle{empty}
\pagestyle{empty}

\begin{abstract}
Autonomous robotic wiping is an important task in various industries, ranging from industrial manufacturing to sanitization in healthcare. Deep reinforcement learning (Deep RL) has emerged as a promising algorithm, however, it often suffers from a high demand for repetitive reward engineering. Instead of relying on manual tuning, we first analyze the convergence of quality-critical robotic wiping, which requires both high-quality wiping and fast task completion, to show the poor convergence of the problem and propose a new bounded reward formulation to make the problem feasible. Then, we further improve the learning process by proposing a novel visual-language model (VLM) based curriculum, which actively monitors the progress and suggests hyperparameter tuning. We demonstrate that the combined method can find a desirable wiping policy on surfaces with various curvatures, frictions, and waypoints, which cannot be learned with the baseline formulation.
The demo of this project can be found at: \url{https://sites.google.com/view/highqualitywiping}
\end{abstract}

\section{INTRODUCTION}

Robotic surface wiping is an important manipulation task with wide domains, such as automation and healthcare. Active research areas involve state detection, trajectory planning, and the low-level interaction skills with surfaces. 
Our work focuses on learning surface interaction skills with a blind policy. A blind wiping policy is often required and cost-effective for scenarios without obstacles, such as wiping tables or car surfaces and handling workpieces. This problem has been commonly approached by classical model-based approaches, which often leverage operational space control and impedance control~\cite{qian2019sensorless, lin2022real, li2022real}, particularly on a flat surface.
However, it is not straightforward to design a model-based controller that works on a variety of surfaces with different curvatures and friction parameters~\cite{amersdorfer2020real,iskandar2023hybrid,vazquez2023continuous}.

Our work investigates learning-based algorithms to take uncertainties into consideration. Unlike traditional approaches that rely on predefined models, learning-based algorithms often demonstrate robust performance in such uncertain environments by leveraging a massive amount of simulation samples. We utilize deep reinforcement learning (deep RL) to generate high-level policies through simulation without prior demonstrations, for dynamic adaptation to complex environmental variables. 
As a result, deep RL will allow us to obtain an autonomous hybrid pose/force controller for precise navigation and force control during wiping tasks 
on surfaces with varying curvatures and friction.

Our research addresses a critical challenge in applying RL to real-world robotic tasks: the inadequacy of off-the-shelf RL approaches for quality-critical tasks.
During RL training, we observed that navigational wiping with quality control is essentially a ``quality-critical'' Markov Decision Process (MDP) problem, demanding a critical balance between fast task execution and high-quality wiping. This duality makes the task very sensitive to hyperparameters. The naive formulation of step-wise rewards for quality instruction and episodic sparse rewards for completion, can easily lead to either degrading work qualities or incentivizing the avoidance of task completion. In fact, the sensitive hyperparameter tuning would be a common issue for many real-world robotic tasks, which has been approached by extensive, labor-intensive manual tuning through repeated trial-and-error.

\begin{figure*}[ht!]
  \centering
  \includegraphics[width=\textwidth]{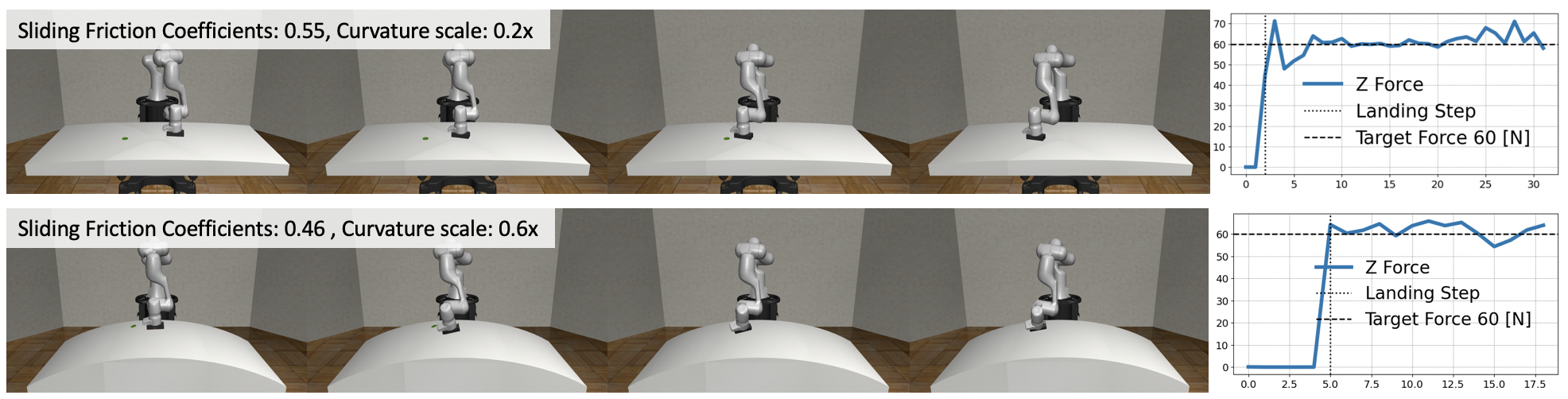}
  \vspace{-2.0em}
  \caption{The example trajectories of the learned wiping policy on surfaces with different curvatures and frictions.
  }
  \vspace{-1.5em}
  \label{fig:force-profile-0914}
\end{figure*}

To address this parameter-sensitive, multi-task learning in RL training, we first demonstrate the infeasibility of the naive formulation, and developed two techniques that we believe are generalizable to tasks facing similar challenges of balancing procedure qualities control and rapid task completion: (i) a bounded reward design with concentric circular checkpoints, which is theoretically grounded, proving that desired behaviors inherently lead to maximal rewards; and (ii) a novel visual-language model (VLM) based curriculum system that simulates human reward engineering, leveraging \replaced{semantic}{multi-modal} understanding and proposing new reward weights. These methods combined, make the convergence supported by thorough analysis while reducing the laborious efforts of fine-tuning required from human researchers.

We show that our novel framework with two novel inventions, bounded reward and VLM-based curriculum, can practically improve the learning process by performing evaluations in a MuJoCo-based environment with variable curvatures, frictions, and waypoint positions. For a 2-points navigation task with a target force of 60N, following 800k training steps, our method yielded a 98\% success rate (+69\%) in navigation, and an average Integral Absolute Error (IAE) of 243 (-9\%), over 25 (-34\%) average completion steps.

\textbf{To summarize, our main contributions are as follows:}
\begin{enumerate}
    \item We formally analyze the convergence of quality-critical robotic wiping and prove the infeasibility\added{ of the naive formulation}.
    \item We propose a new bounded reward function that makes the problem feasible.
    \item We propose a novel VLM-based curriculum for automated and effective parameter tuning.
    \item We demonstrate the effectiveness of the combined learning framework.
\end{enumerate}

\section{Related Work}

\subsection{Robotics Surface Wiping}

Recent works leverage visual observations to generate synthesizing cleaning plans~\cite{hess2011learning, elliott2018robotic}, bounding box and litter classification~\cite{yin2020table}, dense waypoints~\cite{cauli2018autonomous, kim2018icub}, or high-level waypoints with crumbs/spill dynamics modeling~\cite{lew2023robotic}. 

The need for contact force control in robot manipulators, beyond simple position control, is detailed in a survey paper~\cite{suomalainen2022survey} and its references. Several studies utilize dynamic models or sensor feedback for constant contact force and pose correction on unknown curved surfaces~\cite{qian2019sensorless, lin2022real, li2022real, zapico2024semi}. Others use learning-based methods for better generalization to different tools and surfaces. Existing works include learning from demonstration (LfD) and applying motion to different flat, rectangular and horizontal surfaces~\cite{elliott2017learning}; using reinforcement learning~\cite{zhang2020robotic} for tangential angle estimation and constant force tracking; using deep learning network to learn the surface material embedding~\cite{kawaharazuka2022learning}, image embedding of different 3D objects (e.g., cubes, rounds)~\cite{saito2020wiping, saito2021utilization} and subsequent motion control.

\subsection{Deep Reinforcement Learning for Robotics Manipulation}

Deep Reinforcement Learning (DRL) has become pivotal for robotic tasks, complemented by Learning from Demonstration (LfD) which has shown promising outcomes (e.g., \cite{kim2018icub, gams2016adaptation}). Significant progress in robotic manipulation pre-training via demonstrations has been reported \cite{brohan2023rt}. 
Yet RL remains critical for autonomously enhancing simulated demonstrations and subsequent refinement for adaptations.

Our approach diverges in two key aspects from each. Firstly, unlike Zhang et al.~\cite{zhang2020robotic}'s focus on tangential angle estimation and constant force tracking, our emphasis lies on integrating force control within navigational tasks. 
Secondly, unlike Lew et al.~\cite{lew2023robotic} concentrates on crumb collection and spill cleaning on a fixed surface, we train wiping control policies across environments of varying curvatures and smoothness; in contrast to \cite{lew2023robotic}'s use of admittance control with a pre-set normal force, which may falter or prove costly in dynamically changing environments, we gain force control through learning in varied training environments, adaptively determining control inputs.

\subsection{Language to Reward}
Recent efforts have integrated large language models (LLM) with robotics for plan generation, skill bootstrapping, state representation and language-conditioned manipulation. Our work on a visual-language model (VLM) curriculum contributes to the Language to RL Reward initiative, which focuses on converting language into actionable robotic rewards. Notably, EUREKA~\cite{ma2023eureka} automates reward code evolution from environmental and task descriptions through evolutionary optimization based on RL feedback~\cite{ma2023eureka}; TEXT2REWARD~\cite{xie2023text2reward} takes in similar inputs but incorporates human feedback after each RL cycle~\cite{xie2023text2reward}. 
Yu et al.~\cite{yu2023language} uses heuristic templates to transform task descriptions into reward parameters for model predictive control (MPC)~\cite{yu2023language}.

Our VLM-based curriculum can be viewed as an extension of EUREKA~\cite{ma2023eureka} adapted for our learning purpose: Eureka \added{has a LLM agent} update the whole reward function and retrains from scratch for each iteration; we start with a structured RL reward formula to avoid known undesired behaviors, and only update the reward weights during the training process to balance different learning goals. In addition, we add a \replaced{separate vision-language model (VLM) agent}{VLM} to get visual policy replay feedback without extensive logging, analogy to human experiences.

\section{Robotic Wiping as Quality-critical MDP}
We will first formalize the problem of robotic wiping as a common Markov Decision Processes (MDPs) with dense rewards provided per step and sparse reward per episode. Then, we will show the infeasibility of the given wiping task because it is a quality-critical task. Then, we propose a new bounded formulation that makes the problem feasible.

\subsection{Initial Formulation of Markov Decision Process}
We formulate robotic wiping as a Partially Observable Markov Decision Process (POMDP), which is a tuple of the state space $S$, the observation space $O$, the action space $A$, the reward function $r$, the initial state distribution $\rho_0$, the transition function $P(\stp | \st, \at)$, and the discount factor $\gamma$. Our problem is partially observable because certain information, such as the tabletop's curvature and smoothness, is inaccessible due to limited sensory feedback. Then our goal is to find the optimal policy $\pi: {O} \mapsto {A}$ that maximizes the expected episodic reward: $E_{\vc{s}_0 \sim D}[\sum_{t=0}^T \gamma^t r(\st, \at)]$.

\textbf{State:} the state $\vc{s} \in S$ is defined as the internal state of the physics-based simulation.

\textbf{Observation:} A $46$ dimensional observation vector $\vc{o} \in O $ includes waypoint information, joint positions and velocities encoded as sine and cosine of their values, end-effector position and orientation, and force/torque sensor values.

\textbf{Action:} we use a six dimensional pose control to directly adjust the precise position and orientation of the end-effector, which also indirectly adjusts the forces.

\textbf{Reward:}
The reward function $r$ is defined as a weighted sum of the five terms:
\begin{equation}
r(\st, \at, \stp) = 
\begin{cases} 
r_{\text{col}} & \text{if collides} , \\
r_{\text{con}} + r_{\text{force}} + r_{\text{way}} + r_{\text{ac}} & \text{otherwise},
\end{cases}
\label{eq:reward}
\end{equation}
where we omit their arguments for brevity. We also encapsulate all the weights inside of the terms.
If collision happens, agent will receive a negative scalar reward $r_{\text{col}} = -w_{col}$ to penalize collision with the episode terminates immediately. 
Otherwise, we consider four terms that are contact flag, contact force, waypoint, and acceleration penalization rewards.
First, the contact flag reward is defined as $r_{\text{con}} = w_{\text{con}} \mathbf{I}_{\text{con}}$, where $\mathbf{I}_{\text{con}}$ is a zero or one binary flag whether the end effector makes any contact with the table.
The second force term, $r_{\text{force}}$ encourages force control while moving towards the target, which is defined as:
\begin{equation}
r_{\text{force}} = w_{\text{force}} \exp\left( -\frac{(f_z-\mu)^2}{2\sigma^2} \right) \mathbf{I}_{\text{align}},
\label{eq:rt_force}
\end{equation}
Where $w_{\text{force}}$ is the weight, $f_z$ is the upward/downward force applied at the force sensor at EE, and  $e^{-\frac{(f_z-\mu)^2}{2\sigma^2}}$ is a Gaussian shape reward centering at target force $\mu$ (in our case, $\mu =60\text{N}$). 
$\mathbf{I}_{\text{align}}$ is a binary flag which checks the alignment between EE's movement direction and the direction toward the next way point, which returns one when their cosine similarity is greater than 0.8.

The waypoint reward $r_{\text{way}} = w_{\text{way}}\mathbf{I}_{\text{way}}$ denotes the positive episodic reward agent receives for wiping each way point, as $\mathbf{I}_{\text{way}}$ indicates the completion of the waypoint. If the last waypoint is wiped, an extra sparse episodic reward will be provided, end the episode ends. Finally, the term $r_{ac}(\at) := w_{\textbf{ac}}(|a_x| + |a_y| + |a_z|)$ penalizes excessive actions, where $a_x$, $a_y$, $a_z$ are agent's accelerations at $x$, $y$, $z$ axis respectively.

\subsection{Convergence Analysis of Quality-critical MDP} \label{sec:undesired-behavior}

The reward formulation in the previous section consists of common terms in robot learning: dense stepwise feedback to promote desired behaviors and substantial completion rewards to encourage the fast completion. In practice, many researchers typically tune the ratios with many rounds of trial and error to obtain the desirable behaviors. However, tuning hyperparameters for tasks requiring both in-process quality and rapid completion presents significant challenges.

Let us simplify two rewards: a continuous quality reward $W_q$ and an episodic terminal reward for wiping all waypoints, $W_T$. In our case, $W_q$ considers $r_{\text{con}}$, $r_{\text{force}}$, and $r_{\text{ac}}$ while $W_T$ corresponds to the waypoint reward $r_{way}$. We have $W_q^{\text{max}} > 0$ for constant contact with target force and small accelerations, $W_q^{\text{poor}} < W_q^{\text{max}}$ for all other undesired qualities, and $W_T > 0$ to encourage completion. Then, the agent can learn one of three possible strategies, and get respective accumulated rewards:

\begin{itemize}
    \item \textbf{optimal}: takes the best quality wipe and terminates at minimum required time $\text{T}_2$ steps: $\sum_{t=0}^{T_2} \gamma^t W_q^{\text{max}} + \gamma^{T_2} W_T$.
    \item \textbf{lazy}: suboptimal, finishes episode as early as possible without maintaining wiping qualities (e.g., jumping between waypoints with high accelerations and no constant contacts): $\sum_{t=0}^{T_1} \gamma^t W_q^{\text{poor}} + \gamma^{T_1} W_T$.
    \item \textbf{forever}: suboptimal, keeps getting a quality reward without task completion: $\sum_{t=0}^{\infty} \gamma^t W_q^{\text{max}} = W_q^{\text{max}}/(1-\gamma)$.
    
\end{itemize}

For stable learning, it's crucial to establish a feasible relationship between \(W_T\) and \(W_q^{\max}\) so that accumulated rewards meet the constraints for episodes of varying lengths \(\text{T}_1 < \text{T}_2\) and for any \(W_q^{\text{poor}} < W_q^{\text{max}}\). From $\textbf{R}_{\textbf{optimal}} \gg \textbf{R}_{\textbf{lazy}}$, we get the relation below.
\begin{equation}
W_T \ll (\sum_{t=0}^{T_2} \gamma^t W_q^{\text{max}} - \sum_{t=0}^{T_1} \gamma^t W_q^{\text{poor}})/(\gamma^{T_1} - \gamma^{T_2}), 
\label{eq:cond1}
\end{equation}
And from $\textbf{R}_{\textbf{optimal}} \gg \textbf{R}_{\textbf{forever}}$, we get the relation below.
\begin{equation}
W_T \gg (\sum_{t=T_{2+1}}^{\infty} \gamma^t W_q^{\text{max}})/\gamma^{T_2} ,
\label{eq:cond2}
\end{equation}

By combining Eqs.~\ref{eq:cond1} and \ref{eq:cond2}, we want to find a feasible range of $W_T$ regarding $W_q^{\text{max}}$:

\begin{equation}
\text{L}(\text{W}_q^{\text{max}}) \ll W_T \ll \text{U}(\text{W}_q^{\text{max}})
\label{eq:cond3}
\end{equation}

Where $\text{U}(\text{W}_q^{\text{max}}) = (\sum_{t=0}^{T_2} \gamma^t W_q^{\text{max}} - \sum_{t=0}^{T_1} \gamma^t W_q^{\text{poor}})/(\gamma^{T_1} - \gamma^{T_2})$ and $\text{L}(\text{W}_q^{\text{max}}) = (\sum_{t=T_{2+1}}^{\infty} \gamma^t W_q^{\text{max}})/\gamma^{T_2}$.
Finding the lower bound of $\text{U}(\text{W}_q^{\text{max}})$ is more straightforward, as $\text{T}_2/\text{T}_1$ predominantly influences the exponential terms, while $\text{W}_q^{\text{poor}}/\text{W}_q^{\text{max}}$ affects only the linear terms.
We can approximate the lower bound of $\text{U}(\text{W}_q^{\text{max}})$ by setting $\text{T}_1 = 1$ and $\text{T}_2 = \text{H}$, where $\text{H}$ denotes the episode horizon (in our case, $\text{H} = 200$). Then $\text{U}(\text{W}_q^{\text{max}}) \in [99.02\text{W}_q^{\text{max}}, 101.30\text{W}_q^{\text{max}}]$ for $\text{W}_q^{\text{poor}}/\text{W}_q^{\text{max}} \in [0.01, 0.99]$.

On the contrary, finding a feasible $\text{L}(\text{W}_q^{\text{max}})$ applicable for all $\text{T}_2$ is more challenging and prevents the the feasible range of current formulation, which motivates the next section.

\subsection{Bounded Reward Design for Improved Feasibility} \label{sec:bounded-reward-design}

\begin{wrapfigure}{r}{0.4\linewidth}
  \centering
  \includegraphics[width=0.80\linewidth]{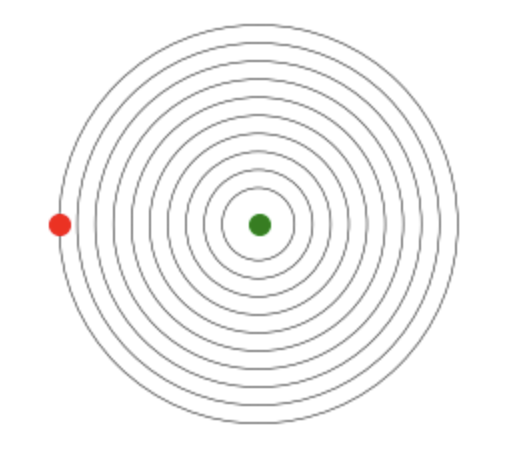}
  \caption{Illustration of Checkpoint Regions.}
  \label{fig:checkpoint}
  \vspace{-1em}
\end{wrapfigure}

To address $\text{L}(\text{W}_q^{\text{max}})$, we introduce concentric circular checkpoint regions between waypoints to promote navigation, inspired by the horizontal checkpoints in the Google research football environment \cite{kurach2020google}. This setup introduces a bounded reward mechanism for target force control $r_{\text{force}}$ and constant contact $r_{\text{con}}$, as outlined in equation~(\ref{eq:reward}). %

Fig.~\ref{fig:checkpoint} illustrates these checkpoint regions around a waypoint, with the next waypoint marked by a green point at the center of equally distanced concentric circles. Rewards $r_{\text{force}}$ and $r_{\text{con}}$ are granted per checkpoint region rather than per step. The updated reward function is similar to equation~(\ref{eq:reward}) but with an additional indicator function:

\begin{equation}
r = 
\begin{cases} 
r_{\text{col}} & \text{if collides} , \\
\mathbf{I}_{\text{check}} (r_{\text{con}} + r_{\text{force}}) + r_{\text{way}} + r_{\text{ac}} & \text{otherwise}.
\end{cases}
\label{eq:reward1}
\end{equation}

Now, two positive terms, $r_{\text{con}}$ and $r_{\text{force}}$ are controlled by the checkpoint indicator $\mathbf{I}_{\text{check}}$, which limits the occurrence of those terms to the number of the checkpoints. This gives us a direct way to bound the cumulative reward $\text{R}_\text{forever}$, which is re-defined from  $\sum_{t=0}^{\infty} \gamma^t W_q^{\text{max}}$ to  $\sum_{t=0}^{T_2} \gamma^t W_{q1}^{\text{max}} + \sum_{t={T_2+1}}^{\infty} \gamma^t W_{q2}^{\text{max}}$, where $\text{T}_2$ is approximated by the time to traverse each checkpoint region only once. $W_{q1}^{\text{max}}$ is identical to $W_{q}^{\text{max}}$ within checkpoint regions, but $W_{q2}^{\text{max}} < 0$ only contains acceleration penalties when all checkpoint regions have been visited. And hence $\text{L}(\text{W}_q^{\text{max}})$ is re-defined as: 

\begin{equation}
\text{L}(\text{W}_q^{\text{max}}) = (\sum_{t={T_2+1}}^{\infty} \gamma^t W_{q2}^{\text{max}})/ \gamma^{T_2}
\label{eq:reward1}
\end{equation}

Given $\text{L}(\text{W}_q^{\text{max}}) \ll 0$, equation~\ref{eq:cond3} holds for $0 < \text{W}_T \ll 99\text{W}_q^{\text{max}}$, altering the policy convergence landscape. Our experiments demonstrate this effectively prevents the convergence to perpetual wiping, as elaborated in Section~\ref{sec:res}.

\

\section{Visual-Language Model based Curriculum} \label{sec:vlm}

\begin{figure*}[t]
  \centering
    \includegraphics[width=0.83\textwidth]{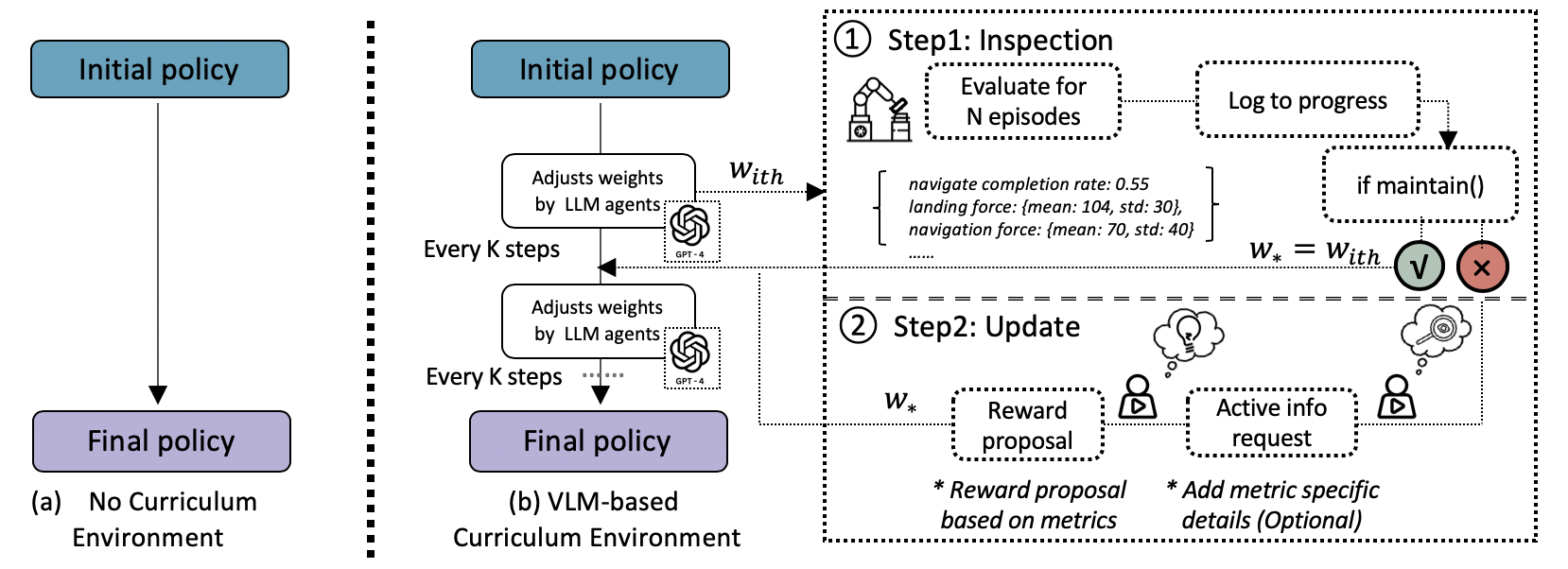}
  \caption{Diagram of the Proposed VLM Algorithms, simulating human decision process on reward scale engineering.
  }
  \label{fig:alog}
  \vspace{-1.5em}
\end{figure*}

While the new formulation makes the quality-critical problem feasible, learning is still hyperparameter sensitive.
To ensure successful trajectories exist and hence can be learned subsequently, we propose a novel Vision-Language Models (VLM) based curriculum learning system, which automatically monitors training metrics and adjusts relative weights of reward terms during the learning process, which resembles the parameter tuning process of human experts.

\subsection{VLM-based Curriculum}

Our learning framework calls the VLM-based curriculum every K steps after the initial M training steps, where K and M are hyper parameters.
The curriculum module auto-adjusts reward weights
for the next cycle with following steps:

\textbf{Step1: Inspection.} In this step, our goal is to collect the initial set of information, which includes success rates, landing pressure profiles, and navigation pressure stats. These stats can be collected by expanding the rollout of the current policy $\pi$ for $N$ times. We maintain the history of the previous information for reference. Once the information is collected, the system checks the pre-defined predicates (e.g., force variance decreased without a significant reduction in navigation success rates) to see if it wants to call the VLM-based hyperparameter tuning.

\textbf{Step2: Update} 
In this step,\added{ there are two large model agents involved: a LLM agent and a VLM agent. } The LLM takes in provided metric \added{from Step 1} and updates reward weights. Depending on the training progress, the LLM could request for different extra information before updating. If the completion rate is low, vision feedback of ending scene summarized by a separate VLM will be provided to describe failure reasons (e.g., no contact, or close to endpoint without finish wiping). If the force metrics require improvements, detailed force percentiles will be provided. This step is desired with multiple purposes: 1) Only providing necessary details into prompts to avoid LLM's catastrophic forgetting on important information. 2) Navigation failures can arise from various scenarios. Leveraging VLM’s semantic capabilities allows us to understand the causes of failures, reducing the need for labor-intensive monitoring and iterative metric development. 3) This hierarchical approach enhances system's extensibility. \added{4) Separating LLM and VLM optimizes reasoning and visual data interpretation respectively.}

\added{The final metrics and extra information will be feed to the LLM. The output from LLM consists of two parts: 1) A 1-2 sentences step-by-step analysis on logs and focus-learning areas; 2) python code for updated reward parameters.}

\added{Detailed prompts can be found at our website noted in the abstract.} The high-level description is summarized in Algorithm~\ref{algo:vlm-curriculum} and Figure~\ref{fig:alog}.

\makeatletter
\algnewcommand{\LineComment}[1]{\Statex \hskip\ALG@thistlm \(\triangleright\) #1}
\algnewcommand{\IndentLineComment}[1]{\Statex \hskip\ALG@tlm \(\triangleright\) #1}
\makeatother

\begin{algorithm}
\caption{VLM-based Curriculum Learning}
\label{algo:vlm-curriculum}
\begin{algorithmic}[1]
\State \textbf{Data:} \added{pre-trained LLM $L$ and }VLM $V$
\State \textbf{Data:} a \added{RL} policy $\pi$
\State \textbf{Data:} a \added{reward weights} parameter vector $\mathbf{w}$
\State $d \gets$ dict(), $i \gets 0$
\While{not converged}
    \State $\pi \gets$ learn($\pi$, $\mathbf{w}$, K) \Comment{Learn a policy for $K$ steps}
    \LineComment{Step 1. Inspection}
    \State $d \gets$ eval($d$, $\pi$, $i$) \Comment{Eval $\pi$ and update $i$th iter data}

    \If{not $\text{maintain}()$}
        \IndentLineComment{Step 2. Update}
        \State $d \gets$ request\_extra\_info\_if\_needed($V$, $d$, $i$)
        \State $\mathbf{w} \gets$ update\_reward\_params(\replaced{$L$}{$V$}, $d$)
    \EndIf
    \State $i += 1$
\EndWhile
\end{algorithmic}
\end{algorithm}
\vspace{-1.5em}

\subsection{Implementation details}

\replaced{We used gpt-4~\cite{achiam2023gpt} as LLM and gpt-4-vision-preview~\cite{openai2023gptv} as VLM.}{We implemented GPT-4~\cite{achiam2023gpt} as the VLM.} To ensure thorough exploration of initial parameters, we initiate our module at 300k steps. We evaluate $N=50$ episodes every $K=100k$ steps and invoke the LLM curriculum module unless evaluation metrics meet the maintenance criteria: an improvement in force profiles—defined by a mean force deviation from the target of less than 5N with reduced variance—without significantly compromising the navigation completion rate (a permissible change of less than 15\%). Initially, $\text{W}_T = 1000$ and $\text{W}_q^{\text{max}} = 29$, which is far from the upper limit of the feasible range $0 < \text{W}_T \ll 99\text{W}_q^{\text{max}}$ outlined in Section \ref{sec:bounded-reward-design}. \added{Optionally, researchers can clip the reward weights for each goal do not exceed twice the weight of any other goal for safeguard.} Throughout the fine-tuning process, the ratio consistently remains within this range.

\section{Results}
We design our experiments to answer the questions below.
\begin{enumerate}
\item Can our methodology effectively train a quality-critical wiping policy for various surfaces?
\item Can (i) Bounded Reward Design, and (ii) VLM-based curriculum improve the learning effectiveness?
\end{enumerate}

\subsection{Experiment Setup}

Our simulation environments are built on top of Mujoco~\cite{todorov2012mujoco} and robosuite~\cite{zhu2020robosuite}. We use the 7-DoF Panda as our robot model, a common choice for both simulated and real-robot research. The trained policies control a robosuite pose controller module using OSC\_POSE option at 20 Hz.

Fig.~\ref{fig:force-profile-0914} illustrates the robot arm performing wiping tasks in various simulated environments. Utilizing domain randomization~\cite{tobin2017domain} for effective Sim2Real Transfer, we generate diverse simulated settings \replaced{randomly sampled at the beginning of each training episode.}{with randomized attributes to train a model effective across all.} Key properties varied include:

\begin{enumerate}
    \item \textbf{Curvature:} Six tabletops with varying curvature (1 flat, 5 curved) to cover a range of surface shapes. The most curved one was created first, and scaled down the z-axis uniformly (flat, 0.2x, 0.4x, 0.6x, 0.8x).
    \item \textbf{Textures:} Sliding ($\mathcal{N}(0.30, 0.05)$), torsional ($\mathcal{N}(0.06, 0.02)$), rolling ($\mathcal{N}(0.0125, 0.005)$) frictions are modeled as Gaussian distributions.
    \item \textbf{Waypoints:} We randomize the location of two waypoints on the tabletop.
\end{enumerate}

We do not include these randomization parameter when we designed curriculum learning.

For analysis, we run experiments for three methods:
\begin{enumerate}
    \item \textbf{non-bounded-reward}: \replaced{The baseline formulation without the bounded reward defined in Fig~\ref{fig:checkpoint}. To balance both objectives, the reward for navigation completion is scaled to match the cumulative wiping quality rewards of the expected completion steps.}{The baseline formulation without the bounded reward defined by checkpoint regions in Fig~\ref{fig:checkpoint}. Rewards are carefully crafted to balance navigation completion and force learning.}    
    \item \textbf{bounded-reward}: The formulation inherits the same reward scales from \textbf{non-bounded-reward}, but incorporated the checkpoint regions as shown in Fig~\ref{fig:checkpoint}.
    \item \textbf{bounded-llm-curr (ours)}: An extended formulation from \textbf{bounded-reward} with VLM-based curriculum discussed in Section~\ref{sec:vlm}. We initialize the learning with the same reward scales, which are \replaced{adjusted by language models during training}{subsequently modified by VLM agents} to enhance learning outcomes. 
\end{enumerate}

\newcommand{\nonbound}{\textbf{non-bounded-reward}\xspace}
\newcommand{\bound}{\textbf{bounded-reward}\xspace}
\newcommand{\llmcurr}{\textbf{bounded-llm-curr}\xspace}
\newcommand{\lowenv}{$\textbf{E}_{\textbf{low}}$\xspace}
\newcommand{\highenv}{$\textbf{E}_{\textbf{enh}}$\xspace}

\subsection{Main Results} \label{sec:res}

\begin{figure*}[ht!]
    \centering
    \begin{subfigure}[t]{0.32\textwidth}
        \centering
        \includegraphics[width=\textwidth]{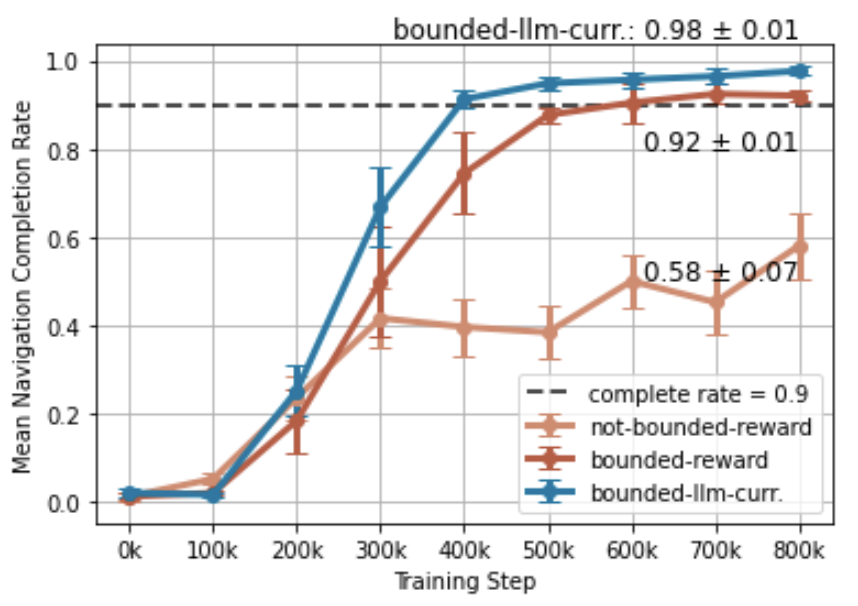}
        \captionsetup{justification=centering}
        \caption{Navigation Complete}
        \label{fig:800k-nav}
    \end{subfigure}
    \begin{subfigure}[t]{0.32\textwidth}
        \centering
        \includegraphics[width=\textwidth]{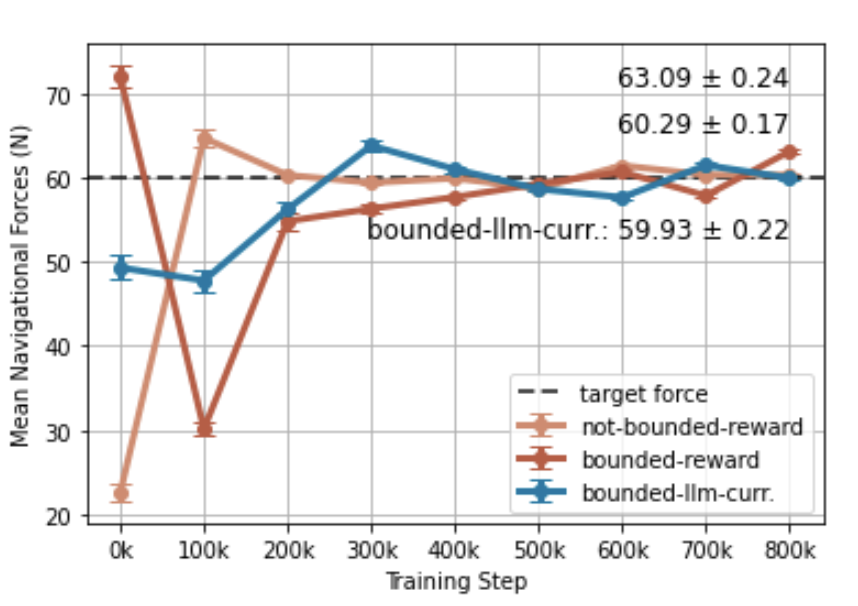}
        \captionsetup{justification=centering}
        \caption{Navigation Force}
        \label{fig:800k-navf}
    \end{subfigure}
    \begin{subfigure}[t]{0.32\textwidth}
        \centering
        \includegraphics[width=\textwidth]{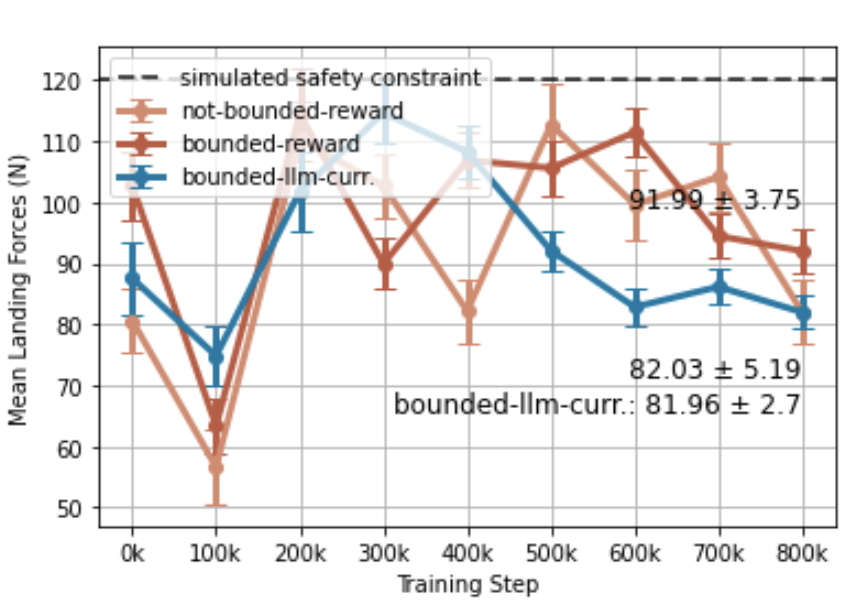}
        \captionsetup{justification=centering}
        \caption{Landing Force}
        \label{fig:800k-landingf}
    \end{subfigure}
    \caption{Evaluation metrics on 2-points environments (line plots with standard error shadows). Force evaluations exclude episodes where the agent wiped repeatedly for the entire horizon without completion -- primarily in the unbounded reward environment -- to mitigate biased distributions. Each method is assessed over 50 episodes with 5 random seeds. }
    \label{fig:progression}
    \vspace{-1em}
\end{figure*}

Our approach effectively trains a wiping policy to navigate waypoints on surfaces with varied curvature and smoothness, while ensuring force remains centered around a target of 60N.
Fig~\ref{fig:progression} demonstrates the successful training outcomes of \llmcurr. It achieves a high navigation completion rate, maintaining stable force control. To illustrate the quality of wiping, we visualize two examples of successful trajectories with different table properties in Fig~\ref{fig:force-profile-0914}, which are nicely centered around our target pressure values $60$N. After 800k steps of training, the policy is able to achieve an average 98\% navigation success rate, and 243 Integral Absolute Error (defined as $\text{IAE} = \int_{0}^{\infty} |e(f)| \, dt,$)\added{ with an average of 25 steps.}

\begin{figure*}[t]
  \centering
    \includegraphics[width=0.83\textwidth]{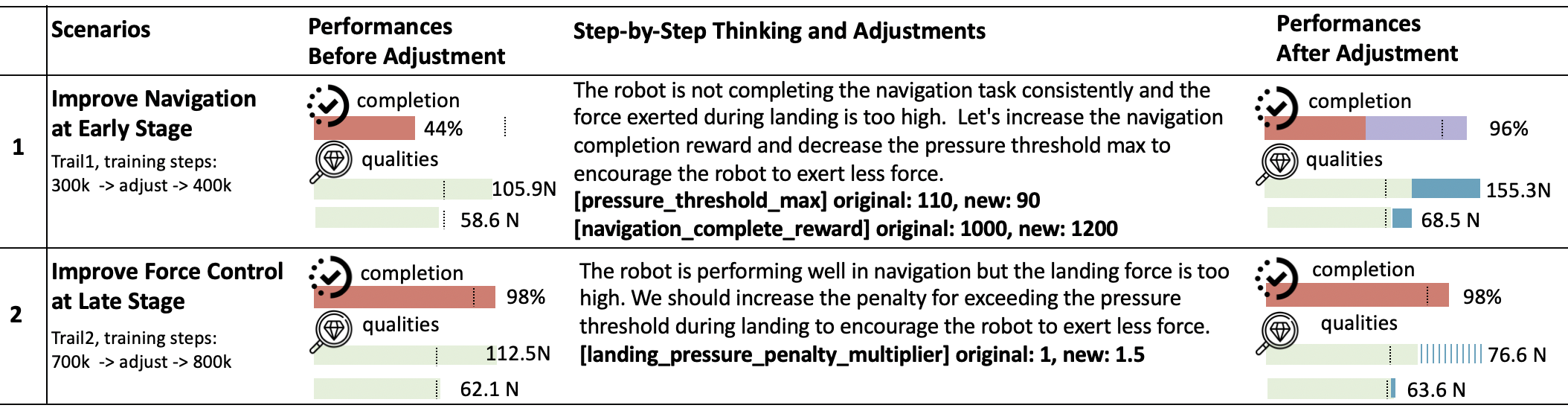}
  \vspace{-0.5em}
  \caption{Examples of VLM-based curriculum adjustment based on the training progresses. Each performance segment includes navigation success rate, average landing pressure (up) and navigational pressure (down). The target pressure is 60N.}
  \label{fig:reason_example}
  \vspace{-1.75em}
\end{figure*}

\begin{table}[H]
    \centering
    \begin{tabular}{l|cccc}
    \hline
    \textbf{Method} & \textbf{Success}  & \textbf{Steps}    & $\textbf{f}_{z}$ IAE  \\ \hline

    \nonbound & 58\% & 38 &  267 \\ %
    \bound & 92\% & 29 & 333 \\ %
    \llmcurr & \textbf{98\%} & \textbf{25} & \textbf{243} \\ 
    
    \hline
    \end{tabular}
    \caption{Evaluation metrics averaged across 5 random seeds. From left to right: navigation success rate; completion steps; IAE of navigational forces.}
    \label{table:metrics}
    \vspace{-1em}
\end{table}

Fig~\ref{fig:800k-nav} and Table~\ref{table:metrics} show the \nonbound method yields around 60\% navigation completion rates, primarily due to suboptimal convergence in four of five seeds, demonstrating as persistent wiping behavior (Section~\ref{sec:undesired-behavior}) in half the cases. We observe no policy converged to such behavior once we introduce bounded reward design as we intended, and hence the navigation success raised significantly from 58\% to 92\%. Further enhancements via a VLM-based curriculum (\llmcurr) increased this rate to 98\%, also optimizing average navigational force accuracy to the target value (60N), reducing Integral Absolute Error ($\text{IAE} = \int_{0}^{\infty} |e(f)|dt, $), shortening completion times, and decreasing landing forces. This strategy effectively trained policies to achieve force control comparable to \nonbound, which prioritized quality at the expense of completion rates, without compromising on the latter.

\subsection{\replaced{Updates and Benefits of VLM Based Curriculum}{Analysis}}

\subsubsection{Efficient fine-tuning with reasoning} 
\added{This section discusses how the system responds to various input metrics and avoids potential local optima for superior solutions, using Fig~\ref{fig:reason_example} as examples. In Scenario 1, when the navigation completion rate is low, the LLM agent increases navigation rewards, enhancing the gradient signals for this metric at the expense of increased landing forces - potentially encouraging successful landings regardless of costs. However, since this occurs early in the training, the RL agent can dedicate the remaining episodes to mastering force control. In Scenario 2, landing force is challenging to learn due to sparse sampling (one per episode). In later training stages, the LLM agent adjusts the penalty multiplier for landing forces, significantly reducing them without adversely affecting other metrics. Combined adjustments lead to better results in Table~\ref{table:metrics}.} To further validate the system, we initiated a set of experiments with imbalanced weight initialization where navigation completion rewards were only 10\% of wiping quality rewards. With \bound, success rates remained near zero even after 600k steps. However, \llmcurr effectively corrected this undesired initialization during the early exploration phase, included successful trajectories, and increased the success rate to 40\% by 500k steps.

\vspace{-0.5em}
\begin{figure}[ht!]
  \centering
  \includegraphics[width=0.42\textwidth]{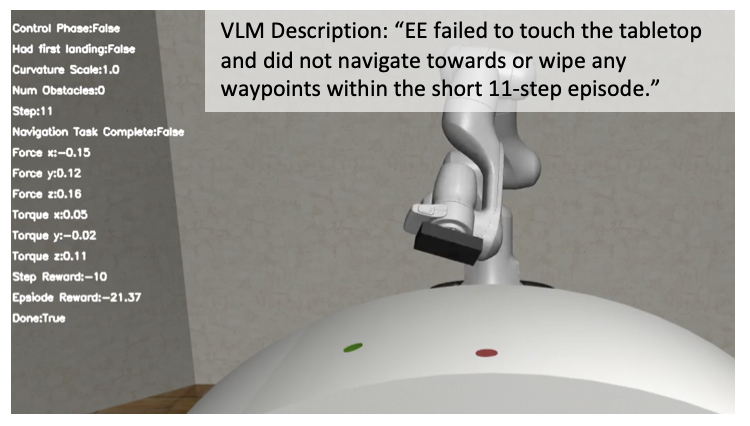}
  \vspace{-0.5em}
  \caption{An example of automatic visual feedback}
  \label{fig:visual}
  \vspace{-1.2em}
\end{figure}

\subsubsection{Visual Monitoring over Failed Behaviors}

\added{Fig~\ref{fig:visual} illustrates how the VLM component summarizes failure reasons. Typically, identifying such open-ended failures requires domain knowledge, iterative monitoring, and extensive logging. In this case, VLM identified the failure occurred early, before contact with the table, leading to a subsequent increase in the intermediate reward for wiping the first waypoint. This example demonstrates the potential of VLMs to enhance understanding in scenarios where the fundamental learning tasks are more complex.}

\section{Conclusion}
This paper presents two techniques for learning effective wiping policies: bounded reward formulation and VLM-based curriculum learning. Initially, we demonstrate the infeasibility of the naive step reward formulation and introduce a bounded approach that improves feasibility. Our novel VLM system actively monitors and adjusts reward weights during learning. Experimental results confirm the efficacy of these methods. We aim to follow up and address current limitations: 1) enhancing the VLM system's generalizability in complex scenarios beyond wiping; 2) deploying policies to hardware to validate real-world performance; and 3) autonomously generating waypoints from observations, thus eliminating the assumption of available waypoints.

\section*{Acknowledgement}
This work was supported by grants from the Electronics and Telecommunications Research Institute (ETRI) [24ZD1130/24BD1300]. We also want to thank Jiachen Yang for his thorough proofreading and insightful feedback.

\bibliographystyle{IEEEtran}
\bibliography{IEEEabrv,references}

\end{document}